\documentclass[11pt]{article}

\usepackage{times}
\usepackage{latexsym}

\usepackage[T1]{fontenc}
\usepackage[utf8]{inputenc}

\usepackage{microtype}

\usepackage{inconsolata}

\usepackage{graphicx}

\pdfoutput=1
\usepackage[final]{acl}

\usepackage{natbib}
\usepackage{times}
\usepackage{latexsym}
\usepackage[T1]{fontenc}
\usepackage[utf8]{inputenc}

\usepackage{microtype}

\usepackage{graphicx} % Package for including graphics
\usepackage{geometry} % Optional: to adjust page margins if needed
\usepackage{inconsolata}
\usepackage{float}
\usepackage{graphicx}
\usepackage{comment}
\usepackage{todonotes}
\usepackage{url}
\title{Simorgh at SemEval-2026 task 7: Region-Aware Hybrid Retrieval for Low-Resource Cultural Reasoning in Multilingual Question Answering}

\author{
Hadi Bayrami Asl Tekanlou$^{1}$ \quad
Mahdi Bakhtiyarzadeh$^{2}$ \quad
Jafar Razmara$^{1}$ \\
$^{1,2}$University of Tabriz, Tabriz, Iran \\
\texttt{h.bayrami1403@ms.tabrizu.ac.ir} \\
\texttt{m.bakhtiyarzadeh1403@ms.tabrizu.ac.ir} \\
\texttt{razmara@tabrizu.ac.ir}
}

\begin{document}
\maketitle
\begin{abstract}
Although Large Language Models (LLMs) demonstrate excellent capabilities and performance for general reasoning tasks within the general public domain, they may face challenges with culturally grounded knowledge within languages with limited digital and textual data. In this paper, we investigate culturally grounded multiple-choice question answering with the BLEnD benchmark, which consists of a multilingual corpus of 30 languages and covers various socio-cultural domains, such as cuisine, sports, family, etc. We propose a region-aware hybrid retrieval approach that combines BM25 lexical matching and dense semantic similarity with regional weighting heuristics to improve the relevance of the answer. The retrieved documents are used to construct a structured prompt for the Qwen3-14B quantized model with logit-based deterministic answer selection. The experimental results show improvements to cross-lingual stability with the hybrid retrieval approach over pure parametric inference for culturally grounded question answering. However, there are still notable performance gaps between languages with more and less training data. This shows that the limitations of the retrieval augmentation approach are not entirely overcome by the training data imbalance problem.
\end{abstract}

\section{Introduction}

Large Language Models (LLMs) have shown impressive capabilities in delivering outstanding results across a broad range of natural language processing tasks ~\cite{schick2023toolformerlanguagemodelsteach,brown2020languagemodelsfewshotlearners}. When asked to answer questions based on Western culture, their answers are highly accurate. However, if they are asked to answer questions related to everyday cultural knowledge from an Eastern background or other less-represented regions, their performance is not as satisfactory ~\cite{Tao_2024,naous2024havingbeerprayermeasuring}. If these models are asked questions related to everyday cultural knowledge, their limitations become quite apparent. For example, if they are asked to answer questions like \textit{“What do Australians need to do to prevent fires in summer?”} their limitations become quite apparent. The reason behind this is that these models are trained on large-scale corpora like Wikipedia and Common Crawl, which are dominated by information from specific regions, languages, and perspectives. The distribution of culturally specific everyday knowledge in these models is inherently unbalanced ~\cite{10.1145/3597307,naous2024havingbeerprayermeasuring}. As a result, these models are likely to produce answers that are incomplete, too general, or even incorrect ~\cite{10319443}. The unbalanced distribution of culturally specific everyday knowledge in these models can result in hallucinations, leading to stereotypical answers ~\cite{myung2025blendbenchmarkllmseveryday}. The limitations of these models in question-answering systems need to be evaluated in relation to everyday knowledge. \\
This research focuses on the answering of multiple-choice questions culturally based with large language models. The following are the main items we will accomplish: 

\begin{itemize}

\item  Create a hybrid retrieval method that incorporates BM25 (what the BM25 algorithm does is attempts to match relevant documents to the query) matched to the query lexically and matched semantically .
    
\item  Implement a region-based enhancement on the retrieval process (where we prioritize evidence that is culturally relevant to the question).
    
\item  Utilize retrieval based augmentation to add culturally appropriate evidence retrieved to the structured prompt for culturally situated reasoning.
    
\item  Apply a logit-based approach for determining the answer choice (A-D) so we can provide deterministic and efficient selection of multiple-choice answers.
\end{itemize}
\section{Related Work}
\subsection{Cultural Commonsense Knowledge in Large Language Models
}
Even though Large Language Models (LLMs) perform well on general types of commonsense reasoning, they do not perform as well when evaluated based on culture-specific commonsense reasoning. The differences between cultures often result in very different performance on the general commonsense tasks provided to LLM's. For example, while both general commonsense and cultural context influence performance of LLMs on commonsense reasoning, the language used in the query modifies the accuracy of the LLM on tasks related to only culture ~\cite{shen-etal-2024-understanding}. These differences in performance indicate that LLM's exhibit built-in biases towards a particular culture, resulting from an unbalanced amount of training data that is heavily weighted toward dominant cultural representations as well as completely monolingual, English-language datasets ~\cite{10.1145/3597307}. The same types of biases exist with respect to moral values. For example, monolingual English LLMs do not adequately account for the fine-grained differences between countries with respect to moral concepts such as homosexuality or divorce, nor do they have a good grasp of the overall patterns of global moral diversity found in datasets such as the World Values Survey or PEW Surveys. The fine-tuning of LLMs using representative data will increase the accuracy of moral inferences across multiple countries but decrease accuracy with respect to norms within the United States ~\cite{ramezani-xu-2023-knowledge}.

\subsection{Performance on Eastern and Region-Specific Cultural Knowledge}
Evaluations of LLMs (Large Language Models) confirm that they perform less well with everyday knowledge of Eastern cultures. In a case study, the Korean medical licensing examination (Korean National Licensing Examination for Korean Medical Doctors, K-NLEKMD) had a 66.18\% successful score for GPT-4. While this was above the cutoff of 60\%, LLMs lost accuracy on localized topics (e.g. public health law, internal medicine, acupuncture) when compared to non-localized topics (compared TKM-specialized versus non-TKM-specialized).Improvements in prompting (e.g., annotating the prompts with Chinese terms, self-consistency) were an important factor in the LLM's successful score; reflecting how LLMs continue to encounter challenges when adapting to other languages and cultures ~\cite{Jang_2023}. n the context of Mandarin Chinese, ChatGPT-3.5 had a passing score of 153.5/300 in the Postgraduate Examination for Clinical Medicine (passed at the 20th percentile), but ChatGPT-3.5's performance on the open-ended exam questions (31.5\% accuracy) failed; results were 42\% in common, 37\% in multi-choice, and 17\% in case analysis, although there was a 90\% average agreement and high insight from the generated results. These results suggest that the recall/diagnosis performance of the LLM was better than that of the intervention-based performance and suggest that the LLM faces challenges adapting to other languages ~\cite{info:doi/10.2196/48514}. Research comparing English to Chinese has shown vast differences between these two languages as well as how LLM's are challenged greatly with Chinese language formality and structure, or grammar, due to cultural differences ~\cite{unknown}.  
\begin{figure*}[h]
    \centering
    \includegraphics[width=1\textwidth]{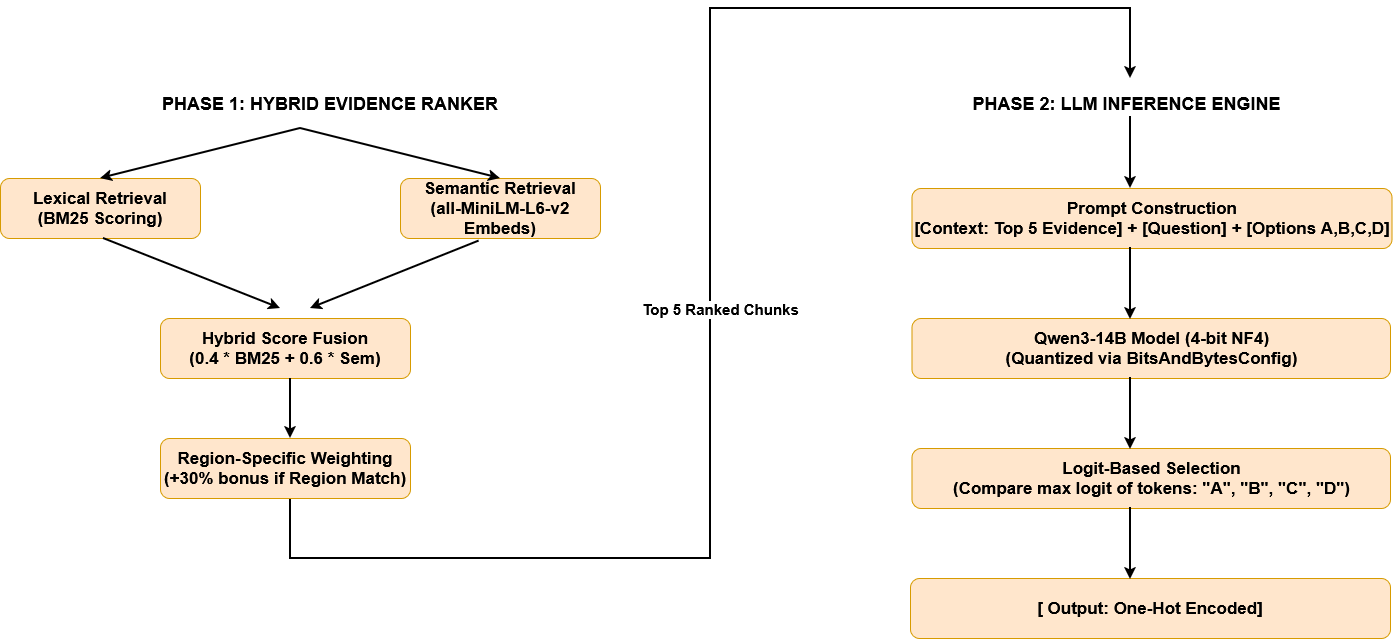}
    \caption{Overview of the proposed system architecture.}
    \label{fig:architecture}
\end{figure*}
\subsection{Bias Origins and Mitigation Strategies}
selection bias of the training data, I.e., it may reflect a variety of social biases (e.g., those related to gender, age, ethnicity, religion, and language) ~\cite{10.1145/3597307}. An imbalanced corpus can also contribute to under-representation of low-resource languages and cultures. Examples of initiatives intended to mitigate this problem include NusaWrites, which creates a culturally relevant corpus using paragraphs written by native speakers in each of the 12 low-resource languages spoken in Indonesia, showing more lexical diversity than scraping or translation methods of obtaining data ~\cite{cahyawijaya-etal-2023-nusawrites}. LLMs present opportunities for researchers in the humanities to use low-resource languages for research; however, researchers face difficulties due to the scarcity of data, adaptability, and cultural sensitivity of the data collected. Researchers require a customized model and collaboration with interdisciplinary teams in order to overcome these obstacles ~\cite{zhong2026opportunitieschallengeslargelanguage}.

\section{Methodology}

We designed a two step pipeline for the BLEnD dataset consisting of:
\begin{itemize}

    \item a Hybrid Evidence Ranker
    \item a Quantized LLM Inference Engine
\end{itemize}
  to address both the cultural rationale of the dataset and the Multiple Choice Question (MCQ) objectives. The overall system architecture is illustrated in Figure~\ref{fig:architecture}. It consists of a hybrid evidence ranking module followed by a quantized LLM-based inference engine for deterministic answer selection.
\subsection{Hybrid Evidence Ranking}
To determine cultural context relevance, our retrieval system provides a combination of traditional lexical matching and dense semantic similarity.
\begin{itemize}
    \item Using the standard BM25 algorithm ~\cite{article} with word-boundary tokenization, lexical matching captures exact matches of keywords between the cultural question and document evidence.
    \item To define contextual nuance and semantic meaning, dense embeddings were produced using the \textit{all-MiniLM-L6-v2} SentenceTransformer model and the cosine similarity between the question and the picture evidence was calculated.
    \item BM25 and semantic scores were combined to obtain an overall score (40\% BM25, 60\% semantic). For further refinement, a heuristic of 30\% weight was added to the combined score if there was a regional reference in the evidence evidence text. The final ranking score for each document is calculated according.
$\mathrm{FinalScore} = (0.4 \cdot \mathrm{BM25} + 0.6 \cdot \mathrm{Semantic}) \cdot (1 + \mathrm{RegionBonus})$
   \item Output: The documents are sorted based on this final score, and the top 5 highest-ranking evidence chunks are extracted and passed to the inference stage.
\end{itemize}

\subsection{LLM-Based Cultural Reasoning}
For the MCQ selection, we utilized a large language model to synthesize the retrieved evidence and determine the correct answer.
\begin{itemize}
    \item Modeling Setup: The Qwen3-14B ~\cite{qwen3technicalreport} model was utilized. To reduce VRAM usage and efficiently batch process models, we loaded the model with 4-bit NormalFloat (nf4) quantization with double quant and using float16 for compute.
    \item Prompt Construction: The model is prompted as a "cultural reasoning model." The model's 5 best documents are inserted directly above the question and choices in the context window, with choices labelled A, B, C and D. If there is no evidence returned, the prompt defaults to using the model's parametric knowledge. 
    \item Prediction Selection based on Logits: Instead of using generated free text, which can have a lot of formatting errors or be very wordy, we used a very strict logarithmic scoring system. The next-token prediction logit values are extracted only for the four letters A, B, C, and D. The choice with the highest probability mass (highest logit) is selected as the final choice.
    \item Prediction Processing and Exporting: Predictions are processed in batches of 16 for maximum use of GPU. The final predictions will be converted to one-hot encoding format for the four choices. 
\end{itemize}
\section{Results}
\subsection{Dataset}
\begin{figure}[h]
    \centering
    \includegraphics[width=0.4\textwidth]{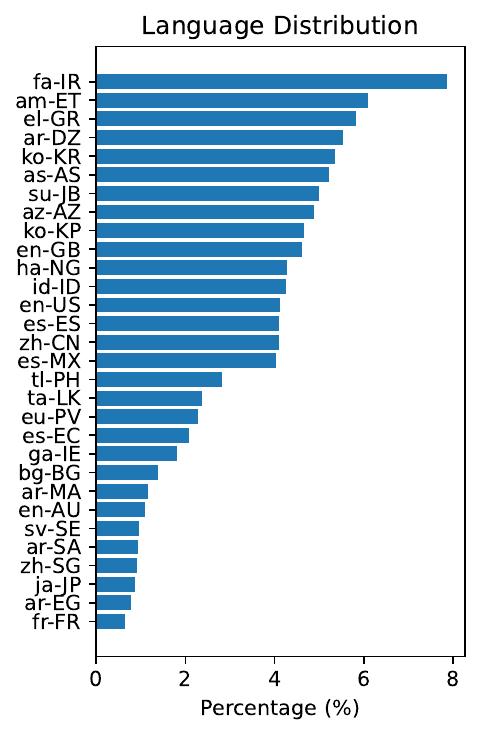}
    \caption{Percentage distribution of question–answer pairs across languages in the BLEnD benchmark.}
    \label{fig:plot1}
\end{figure}

For evaluations, we use the BLEnD benchmark~\cite{myung2024blend}, developed as part of SemEval-2026 Task 7~\cite{semeval2026task7}, which focuses on evaluating everyday knowledge across diverse languages and cultural contexts. It comprises approximately 52,600 question–answer pairs spanning 16 diverse countries/regions and 30 languages, including several low-resource languages such as Amharic, Assamese, Azerbaijani, Hausa, and Sundanese. This diversity ensures a rigorous test of a model's ability to reason across both Western-centric and underrepresented cultural contexts. The questions are hand-crafted by native speakers and categorized into six core socio-cultural domains: Food, Sports, Family, Education, Holidays \& Leisure, and Work-Life. Figure~\ref{fig:plot1} presents the language-wise percentage distribution of question–answer pairs in the benchmark. Although the dataset is multilingual, the representation is not uniform across languages, which may influence cross-lingual evaluation dynamics.
\subsection{Quantitative Results}
To quantify the impact of region-aware hybrid retrieval on culturally situated reasoning, we perform a comprehensive evaluation on the BLEnD benchmark. The most accurate answer (of the four) will be identified through an exact match to determine how accurately the cultural appropriateness of a model's response can be without generating free-text output. As illustrated in Figure 2 ~\ref{fig:plot1}, a significant amount of variance exists in performance levels among different languages; that is, high-resource languages demonstrate much greater accuracy than many low-resource languages. The continued performance disparity between the two groups of languages relates to the disparity of training data available for the models; thus, retrieval-based augmentation affects the degree of performance disparity to a certain extent. However, the overall performance difference is smaller than typically observed in analyses of pure parametric (i.e., non-hybrid) LLMs, suggesting that the hybrid ranking approach may help mitigate some of the existing disparities in cultural knowledge representation. 
\begin{figure*}[h]
    \centering
    \includegraphics[width=1\textwidth]{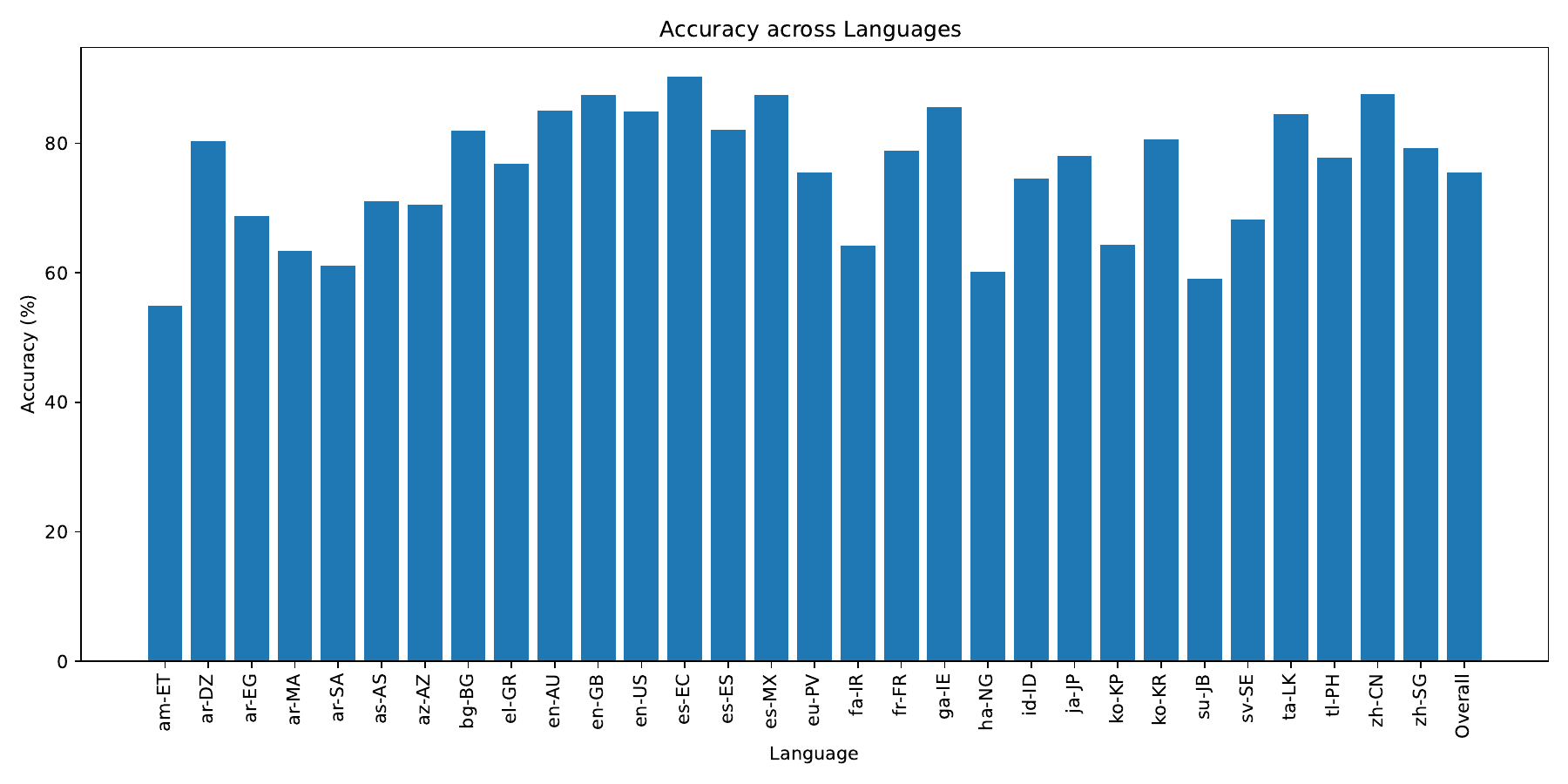}
    \caption{Test-set performance comparison across languages.}
    \label{fig:plot2}
\end{figure*}
\section{Limitation and Conclusion}
Figure ~\ref{fig:plot2}'s quantitative evaluation results indicate that, although the model performs adequately from an accuracy perspective, there is still a substantial lack of success in low-resource languages (e.g., low accuracy). Although the hybrid retrieval mechanism proposed in this study alleviates some of the cultural knowledge gaps present in certain languages, the performance gap remains in languages that were not as well represented during large-scale pretraining (e.g., Amharic, Hausa, and Sundanese) based on the BLEnD benchmark. Therefore, it appears that retrieval augmentation alone will not adequately address some of the imbalances created by using large-scale pretraining corpora for creating a machine translation system. 
One limiting factor is how well the search engine depends on only external document collection. If there are not enough relevant cultural documents available, or if they do not match the query, the retrieval module will be more likely to retrieve irrelevant documents than it will to provide a basis for using meaningful contextual evidence. In addition, the region-weighting heuristic may not work properly in situations where the cultural reference is not stated explicitly, but only found on the surface of a region. Therefore, it may not be effective in many cases because it evaluates cultural documents based on their general location rather than on their specific nature.
A second limitation is that a logit-based deterministic answer selection strategy limits the reasoning to next-token probability distributions and does not model multi-step deliberation directly, even though this provides for robustness of format and computation efficiency. As a result, more nuanced, culturally-grounded reasoning involving deeper inferences may be missed.
Afterwards, 4 bit quantizing allows quick and effective uses of synthetic intelligence, but can result in small losses of representational precision, which can impact the ability of the AI to perform well on languages that are very linguistically complex or grammatically rich. In this paper, we proposed an interesting hybrid retrieval and logit-based inference structure that was created using regionally contextualized data to answer multiple-choice questions with cultural background. Results from our experiments using the BLEnD benchmark indicated that actually incorporating lexical matching, semantic similarity, and regionally prioritized ranking into a hybrid retrieval and logit-based inference model helped stabilize cross-cultural reasoning performance. Nevertheless, the ongoing performance gaps between low resource languages remind us that dissimilar training sets have an ongoing negative impact on the performance of AI systems; and there is a continuing need to use more culturally diverse training and retrieval methods on the development and improvement of AI systems. It is essential for future work to focus on developing dynamic region models, creating a culturally informed re-ranking mechanism, and developing fine-tuning strategies that target low-resource linguistic communities. By enhancing multilingual evidence bases and using structured cultural knowledge graphs to bolster the cross-cultural evaluation of LLMs, we can improve the integrity and fairness of the evaluations.

\bibliography{refs}

@misc{myung2025blendbenchmarkllmseveryday,
      title={BLEnD: A Benchmark for LLMs on Everyday Knowledge in Diverse Cultures and Languages}, 
      author={Junho Myung and Nayeon Lee and Yi Zhou and Jiho Jin and Rifki Afina Putri and Dimosthenis Antypas and Hsuvas Borkakoty and Eunsu Kim and Carla Perez-Almendros and Abinew Ali Ayele and Víctor Gutiérrez-Basulto and Yazmín Ibáñez-García and Hwaran Lee and Shamsuddeen Hassan Muhammad and Kiwoong Park and Anar Sabuhi Rzayev and Nina White and Seid Muhie Yimam and Mohammad Taher Pilehvar and Nedjma Ousidhoum and Jose Camacho-Collados and Alice Oh},
      year={2025},
      eprint={2406.09948},
      archivePrefix={arXiv},
      primaryClass={cs.CL},
      url={https://arxiv.org/abs/2406.09948}, 
}

@misc{brown2020languagemodelsfewshotlearners,
      title={Language Models are Few-Shot Learners}, 
      author={Tom B. Brown and Benjamin Mann and Nick Ryder and Melanie Subbiah and Jared Kaplan and Prafulla Dhariwal and Arvind Neelakantan and Pranav Shyam and Girish Sastry and Amanda Askell and Sandhini Agarwal and Ariel Herbert-Voss and Gretchen Krueger and Tom Henighan and Rewon Child and Aditya Ramesh and Daniel M. Ziegler and Jeffrey Wu and Clemens Winter and Christopher Hesse and Mark Chen and Eric Sigler and Mateusz Litwin and Scott Gray and Benjamin Chess and Jack Clark and Christopher Berner and Sam McCandlish and Alec Radford and Ilya Sutskever and Dario Amodei},
      year={2020},
      eprint={2005.14165},
      archivePrefix={arXiv},
      primaryClass={cs.CL},
      url={https://arxiv.org/abs/2005.14165}, 
}

@misc{schick2023toolformerlanguagemodelsteach,
      title={Toolformer: Language Models Can Teach Themselves to Use Tools}, 
      author={Timo Schick and Jane Dwivedi-Yu and Roberto Dessì and Roberta Raileanu and Maria Lomeli and Luke Zettlemoyer and Nicola Cancedda and Thomas Scialom},
      year={2023},
      eprint={2302.04761},
      archivePrefix={arXiv},
      primaryClass={cs.CL},
      url={https://arxiv.org/abs/2302.04761}, 
}

@article{Tao_2024,
   title={Cultural bias and cultural alignment of large language models},
   volume={3},
   ISSN={2752-6542},
   url={http://dx.doi.org/10.1093/pnasnexus/pgae346},
   DOI={10.1093/pnasnexus/pgae346},
   number={9},
   journal={PNAS Nexus},
   publisher={Oxford University Press (OUP)},
   author={Tao, Yan and Viberg, Olga and Baker, Ryan S and Kizilcec, René F},
   editor={Muthukrishna, Michael},
   year={2024},
   month=sep }

@misc{naous2024havingbeerprayermeasuring,
      title={Having Beer after Prayer? Measuring Cultural Bias in Large Language Models}, 
      author={Tarek Naous and Michael J. Ryan and Alan Ritter and Wei Xu},
      year={2024},
      eprint={2305.14456},
      archivePrefix={arXiv},
      primaryClass={cs.CL},
      url={https://arxiv.org/abs/2305.14456}, 
}

@article{10.1145/3597307,
author = {Navigli, Roberto and Conia, Simone and Ross, Bj\"{o}rn},
title = {Biases in Large Language Models: Origins, Inventory, and Discussion},
year = {2023},
issue_date = {June 2023},
publisher = {Association for Computing Machinery},
address = {New York, NY, USA},
volume = {15},
number = {2},
issn = {1936-1955},
url = {https://doi.org/10.1145/3597307},
doi = {10.1145/3597307},
journal = {J. Data and Information Quality},
month = jun,
articleno = {10},
numpages = {21}
}

@ARTICLE{10319443,
  author={McIntosh, Timothy R. and Liu, Tong and Susnjak, Teo and Watters, Paul and Ng, Alex and Halgamuge, Malka N.},
  journal={IEEE Transactions on Artificial Intelligence}, 
  title={A Culturally Sensitive Test to Evaluate Nuanced GPT Hallucination}, 
  year={2024},
  volume={5},
  number={6},
  pages={2739-2751},
  doi={10.1109/TAI.2023.3332837}}

@inproceedings{shen-etal-2024-understanding,
    title = "Understanding the Capabilities and Limitations of Large Language Models for Cultural Commonsense",
    author = "Shen, Siqi  and
      Logeswaran, Lajanugen  and
      Lee, Moontae  and
      Lee, Honglak  and
      Poria, Soujanya  and
      Mihalcea, Rada",
    editor = "Duh, Kevin  and
      Gomez, Helena  and
      Bethard, Steven",
    booktitle = "Proceedings of the 2024 Conference of the North American Chapter of the Association for Computational Linguistics: Human Language Technologies (Volume 1: Long Papers)",
    month = jun,
    year = "2024",
    address = "Mexico City, Mexico",
    publisher = "Association for Computational Linguistics",
    url = "https://aclanthology.org/2024.naacl-long.316/",
    doi = "10.18653/v1/2024.naacl-long.316",
    pages = "5668--5680"
}

@inproceedings{ramezani-xu-2023-knowledge,
    title = "Knowledge of cultural moral norms in large language models",
    author = "Ramezani, Aida  and
      Xu, Yang",
    editor = "Rogers, Anna  and
      Boyd-Graber, Jordan  and
      Okazaki, Naoaki",
    booktitle = "Proceedings of the 61st Annual Meeting of the Association for Computational Linguistics (Volume 1: Long Papers)",
    month = jul,
    year = "2023",
    address = "Toronto, Canada",
    publisher = "Association for Computational Linguistics",
    url = "https://aclanthology.org/2023.acl-long.26/",
    doi = "10.18653/v1/2023.acl-long.26",
    pages = "428--446"
}

@article{Jang_2023,
   title={GPT-4 can pass the Korean National Licensing Examination for Korean Medicine Doctors},
   volume={2},
   ISSN={2767-3170},
   url={http://dx.doi.org/10.1371/journal.pdig.0000416},
   DOI={10.1371/journal.pdig.0000416},
   number={12},
   journal={PLOS Digital Health},
   publisher={Public Library of Science (PLoS)},
   author={Jang, Dongyeop and Yun, Tae-Rim and Lee, Choong-Yeol and Kwon, Young-Kyu and Kim, Chang-Eop},
   editor={Nakayama, Luis Filipe},
   year={2023},
   month=dec, pages={e0000416} }

@Article{info:doi/10.2196/48514,
author="Yu, Peng
and Fang, Changchang
and Liu, Xiaolin
and Fu, Wanying
and Ling, Jitao
and Yan, Zhiwei
and Jiang, Yuan
and Cao, Zhengyu
and Wu, Maoxiong
and Chen, Zhiteng
and Zhu, Wengen
and Zhang, Yuling
and Abudukeremu, Ayiguli
and Wang, Yue
and Liu, Xiao
and Wang, Jingfeng",
title="Performance of ChatGPT on the Chinese Postgraduate Examination for Clinical Medicine: Survey Study",
journal="JMIR Med Educ",
year="2024",
month="Feb",
day="9",
volume="10",
pages="e48514",
issn="2369-3762",
doi="10.2196/48514",
url="https://mededu.jmir.org/2024/1/e48514",
url="https://doi.org/10.2196/48514",
url="http://www.ncbi.nlm.nih.gov/pubmed/38335017"
}

@unknown{unknown,
author = {Wang, Shiyu and Ouyang, Qian and Wang, Bing},
year = {2024},
month = {02},
pages = {},
title = {Comparative Evaluation of Commercial Large Language Models on PromptBench: An English and Chinese Perspective},
doi = {10.21203/rs.3.rs-3987793/v1}
}

@inproceedings{cahyawijaya-etal-2023-nusawrites,
    title = "{N}usa{W}rites: Constructing High-Quality Corpora for Underrepresented and Extremely Low-Resource Languages",
    author = "Cahyawijaya, Samuel  and
      Lovenia, Holy  and
      Koto, Fajri  and
      Adhista, Dea  and
      Dave, Emmanuel  and
      Oktavianti, Sarah  and
      Akbar, Salsabil  and
      Lee, Jhonson  and
      Shadieq, Nuur  and
      Cenggoro, Tjeng Wawan  and
      Linuwih, Hanung  and
      Wilie, Bryan  and
      Muridan, Galih  and
      Winata, Genta  and
      Moeljadi, David  and
      Aji, Alham Fikri  and
      Purwarianti, Ayu  and
      Fung, Pascale",
    editor = "Park, Jong C.  and
      Arase, Yuki  and
      Hu, Baotian  and
      Lu, Wei  and
      Wijaya, Derry  and
      Purwarianti, Ayu  and
      Krisnadhi, Adila Alfa",
    booktitle = "Proceedings of the 13th International Joint Conference on Natural Language Processing and the 3rd Conference of the Asia-Pacific Chapter of the Association for Computational Linguistics (Volume 1: Long Papers)",
    month = nov,
    year = "2023",
    address = "Nusa Dua, Bali",
    publisher = "Association for Computational Linguistics",
    url = "https://aclanthology.org/2023.ijcnlp-main.60/",
    doi = "10.18653/v1/2023.ijcnlp-main.60",
    pages = "921--945"
}

@misc{zhong2026opportunitieschallengeslargelanguage,
      title={Opportunities and Challenges of Large Language Models for Low-Resource Languages in Humanities Research}, 
      author={Tianyang Zhong and Zhenyuan Yang and Zhengliang Liu and Ruidong Zhang and Weihang You and Yiheng Liu and Haiyang Sun and Yi Pan and Yiwei Li and Yifan Zhou and Hanqi Jiang and Junhao Chen and Tianming Liu},
      year={2026},
      eprint={2412.04497},
      archivePrefix={arXiv},
      primaryClass={cs.CL},
      url={https://arxiv.org/abs/2412.04497}, 
}

@article{article,
author = {Robertson, Stephen and Zaragoza, Hugo},
year = {2009},
month = {01},
pages = {333-389},
title = {The Probabilistic Relevance Framework: BM25 and Beyond},
volume = {3},
journal = {Foundations and Trends in Information Retrieval},
doi = {10.1561/1500000019}
}

@misc{qwen3technicalreport,
      title={Qwen3 Technical Report}, 
      author={Qwen Team},
      year={2025},
      eprint={2505.09388},
      archivePrefix={arXiv},
      primaryClass={cs.CL},
      url={https://arxiv.org/abs/2505.09388}, 
}

@article{myung2024blend,
title={Blend: A benchmark for llms on everyday knowledge in diverse cultures and languages},
author={Myung, Junho and Lee, Nayeon and Zhou, Yi and Jin, Jiho and Putri, Rifki and Antypas, Dimosthenis and Borkakoty, Hsuvas and Kim, Eunsu and Perez-Almendros, Carla and Ayele, Abinew Ali and others},
journal={Advances in Neural Information Processing Systems},
volume={37},
pages={78104--78146},
year={2024}
}

@inproceedings{semeval2026task7,
title = "{S}em{E}val-2026 {T}ask 7: {E}veryday {K}nowledge {A}cross {D}iverse {L}anguages and {C}ultures",
author = "Nedjma Ousidhoum and Junho Myung and Carla Perez-Almendros and Jiho Jin and
Amr Keleg and Meriem Beloucif and Yi Zhou and Rodrigo Agerri and Vladimir Araujo and
Naomi Baes and James Barry and Joanne Boisson and Nancy F. Chen and Christine de Kock and
Aleksandra Edwards and Joseba Fernandez de Landa and Mohamed Fazli Imam and Huda Hakami and
Shu-Kai Hsieh and Joseph Marvin Imperial and Roy Ka-Wei Lee and Chenyang Lyu and
Younes Samih and Johan Sjons and Bryan Tan and Asahi Ushio and Weihua Zheng and Zhengyuan Liu and
Alice Oh and Jose Camacho-Collados",
booktitle = "Proceedings of the 20th International Workshop on Semantic Evaluation (SemEval-2026)",
year = "2026",
publisher = "Association for Computational Linguistics"
}

\end{document}